\pdfoutput=1

\documentclass[11pt]{article}

\usepackage[]{ACL2023}

\usepackage{times}
\usepackage{latexsym}

\usepackage[T1]{fontenc}

\usepackage[utf8]{inputenc}

\usepackage{microtype}

\usepackage{inconsolata}
\usepackage[utf8]{inputenc}
\usepackage{microtype}
\usepackage{subfigure}
\usepackage{svg}
\usepackage{CJKutf8}
\usepackage{setspace}
\usepackage{multirow}
\usepackage{subfloat}
\usepackage{graphicx}
\usepackage{url}
\usepackage{tikz}
\usepackage{pgfplots}
\usepackage{color}
\usepackage{mathtools}
\usepackage{makecell}
\usepackage{colortbl,booktabs}
\usepackage{paralist}
\usepackage{amsmath,amssymb,bm}
\usepackage{threeparttable}
\usepackage{arydshln}
\usepackage{soul}
\usepackage{changes}
\usepackage{bbding}

\title{Improved Visual Story Generation with Adaptive Context Modeling}

\author{Zhangyin Feng \textsuperscript{\rm 1}, 
Yuchen Ren \textsuperscript{\rm 2}, 
Xinmiao Yu \textsuperscript{\rm 1}, \\
\textbf{Xiaocheng Feng} \textsuperscript{\rm 1,3},
\textbf{Duyu Tang}, 
\textbf{Shuming Shi}, 
\textbf{Bing Qin} \textsuperscript{\rm 1,3} \\
$^1$ Harbin Institute of Technology, \\
$^2$ Renmin University of China, \\
$^3$ Peng Cheng Laboratory \\
\{zyfeng, xmyu, xcfeng, qinb\}@ir.hit.edu.cn \\
siriusren@ruc.edu.cn
}
\begin{document}
\maketitle
\begin{abstract}

Diffusion models developed on top of powerful text-to-image generation models like Stable Diffusion achieve remarkable success in visual story generation. However, the best-performing approach considers historically generated results as flattened memory cells, ignoring the fact that not all preceding images contribute equally to the generation of the characters and scenes at the current stage. To address this, we present a simple method that improves the leading system with adaptive context modeling, which is not only incorporated in the encoder but also adopted as additional guidance in the sampling stage to boost the global consistency of the generated story. We evaluate our model on PororoSV and FlintstonesSV datasets and show that our approach achieves state-of-the-art FID scores on both story visualization and continuation scenarios.
We conduct detailed model analysis and show that our model excels at generating semantically consistent images for stories.

\end{abstract}

\section{Introduction}
Diffusion models trained on broad text-image data \cite{latentdiffusion,bao2022all,feng2022ernie,dalle-2,imagen,balaji2022ediffi,nichol2021glide} achieved remarkable success in text-to-image generation and showed strong abilities to synthesize photorealistic images of high resolution and great semantic consistency to text prompts. 
Such a huge success drives the extension of modern diffusion text-to-image models into more scenarios like visual story generation, which is to generate a series of images for a story of multiple sentences.

\begin{figure}[t]
  \includegraphics[width=1\linewidth]{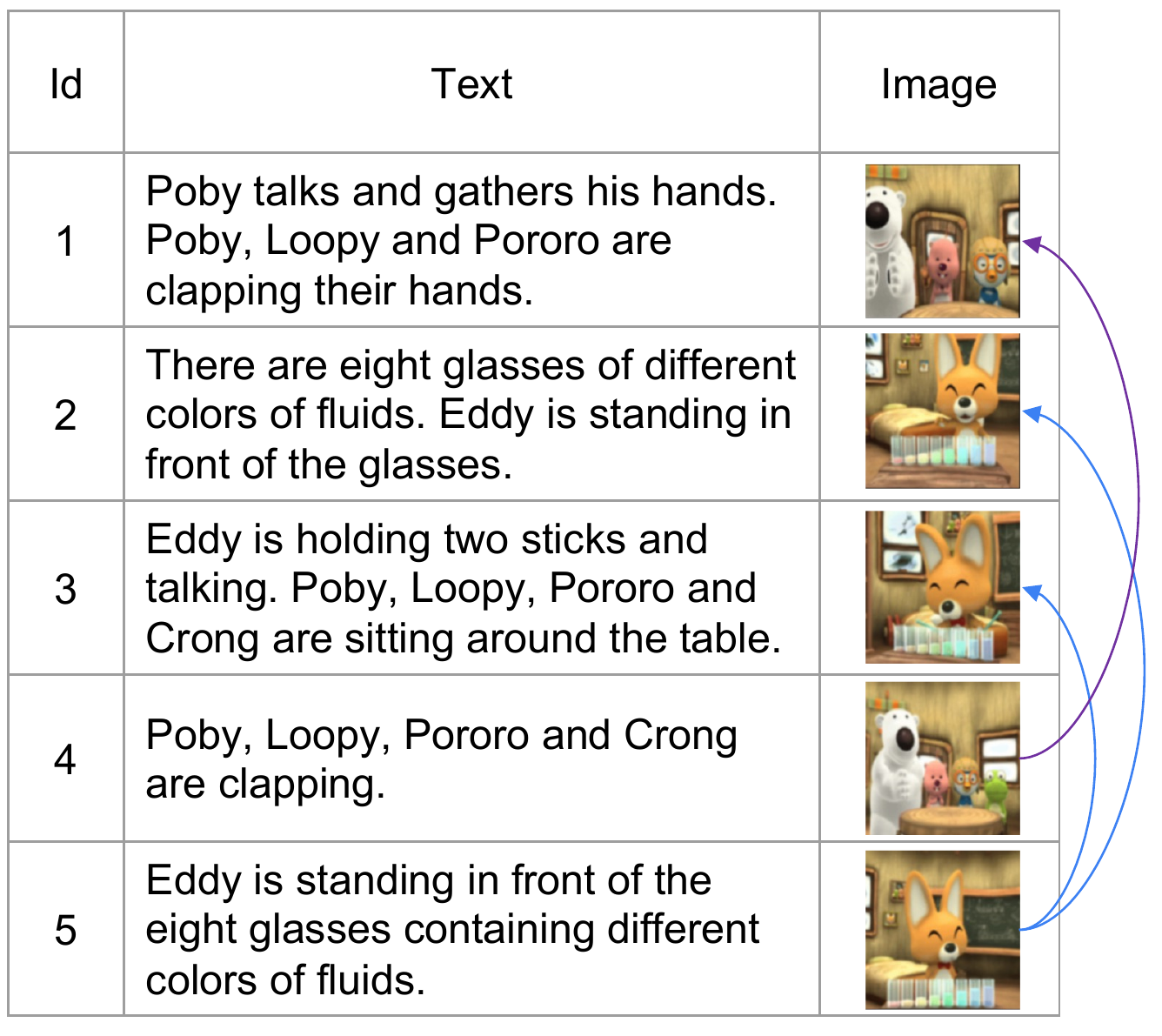}
  \caption{A motivating example of a story with five sentences. Blue and purple lines indicate the dependencies between images.}
  \label{fig:intro}
\end{figure}

A recent work, AR-LDM \cite{pan2022synthesizing}, which is built upon open-sourced Stable Diffusion, achieves the state-of-the-art FID on the benchmark datasets for visual story generation.
AR-LDM encodes previous text-image context as a sequence of additional conditions, which is then attended by the UNet decoder for image generation.
Despite its remarkable success, one limitation is that previous text-image pairs of the same story are all flattened as conditioning memories. This is different from the fact that not all the scenes/characters of sentences in the same story are closely related. Take Figure \ref{fig:intro} as an example. The scene of the fourth sentence is not related to either the second or the third sentence. On the contrary, the generation of the fifth image should depend more on the second and third images than others.
From this example, we can see that the dependency between images could be largely measured by the semantic relations between sentences.

In this work, we present a simple approach \footnote{We name our model as \textbf{ACM-VSG} (\textbf{A}daptive \textbf{Context} \textbf{M}odeling for \textbf{V}isual \textbf{S}tory \textbf{G}eneration).} that selectively adopts historical text-image data from the same story in the generation of an image. Specifically, we freeze the text and image representations produced by off-the-shell encoders, and adaptively compute conditioning vectors of context by considering the semantic relation between the current sentence and all the history. Such resulting conditioning vectors will be queried by UNet in a traditional way. Furthermore, based on the consideration that images should have similar scenes and characters if their corresponding sentences are similar, we further add context-aware guidance like the use of classifier guidance or CLIP guidance \cite{nichol2021glide} in standard text-to-image generation. 

To validate the effectiveness of our approach, we evaluate our model on story visualization and continuation tasks.
Experimental results on PororoSV and FlintstonesSV datasets show that both adaptive encoder and guidance improve the quality of the generated images as well as the global consistency of the visual story.
The contributions of this work are as follows:
\begin{itemize}
    \item We present a diffusion model that adaptively uses context information in the encoder and sampling guidance for visual story generation.
    \item Our approach achieves state-of-the-art results on benchmark datasets for both story visualization and continuation tasks.
    \item We show that our model excels at generating semantically consistent images for stories.
\end{itemize}

\section{Related work}
\subsection{Text-to-Image Generation}

We group modern text-to-image generation approaches into three categories. 
The \textbf{first} category is generative adversarial network \cite{goodfellow2014generative,reed2016generative,zhang2017stackgan}. They jointly learn a generator and a discriminator, where the generator is trained to generate images to fool the discriminator and the discriminator is trained to distinguish between real and (generated) fake images. 
The \textbf{second} category is encoder-decoder plus discrete variational autoencoder (dVAE). Methods are developed based on a well-trained discrete variational autoencoder \cite{van2017dvae}, which is capable of mapping an image to discrete tokens and reconstructing an image from discrete tokens.
Thus, the task of text-to-image generation could be viewed as a special translation task that converts natural language tokens to image tokens. 
Autoregressive models \cite{dalle,ding2021cogview,gafni2022make,parti} typically use Transformer \cite{vaswani2017attention} to generate a visual token conditioned on the previously generated tokens, resulting in high latency in the inference stage. 
Muse \cite{chang2023muse} is a non-autoregressive model that tremendously speeds up the inference stage by generating image tokens in parallel.
The \textbf{third} category is diffusion models \textemdash \  image generation is considered as an iterative refinement process, where two ends of the spectrum are the Gaussian noise and the real image, respectively. 
Some studies adopt a variational autoencoder to compress an image to the latent space and learn the diffusion process in the latent space of images \cite{latentdiffusion,bao2022all,feng2022ernie}. 
Some works \cite{dalle-2,imagen,balaji2022ediffi} directly learn the diffusion model over pixels and typically include cascaded up-sampling models (e.g., from 64$\times$64 to 256$\times$256 and from 256$\times$256 to 1024$\times$1024) to produce high-resolution images. 

\subsection{Visual Story Generation}
Visual story generation includes two settings: story visualization and story continuation.
Story visualization was firstly introduced by \citet{li2019storygan}, who proposes the StoryGAN model for sequential text-to-image generation. Based on the GAN network, they proposed to combine image and story discriminators for adversarial learning.
To improve the global consistency across dynamic scenes and characters in the story, \citet{zeng2019pororogan} jointly considers story-to-image-sequence, sentence-to-image, and word-to-image-patch alignment by proposing an aligned sentence encoder and attentional word encoder.
\citet{li2020improved} includes dilated convolution in the discriminators to expand the receptive field of the convolution kernel in the feature maps and weighted activation degree to provide a robust evaluation between images and stories.
To improve the visual quality, coherence and relevance of generated images, 
\citet{maharana-etal-2021-improving} extends the GAN structure by including a dual learning framework that utilizes video captioning to reinforce the semantic alignment between the story and generated images, and a copy-transform mechanism for sequentially consistent story visualization.
\citet{maharana-bansal-2021-integrating} improves the generation quality by incorporating constituency parse trees, commonsense knowledge, and visual structure via bounding boxes and dense captioning.
\begin{figure*}[h]
 \centering
  \includegraphics[width=\linewidth]{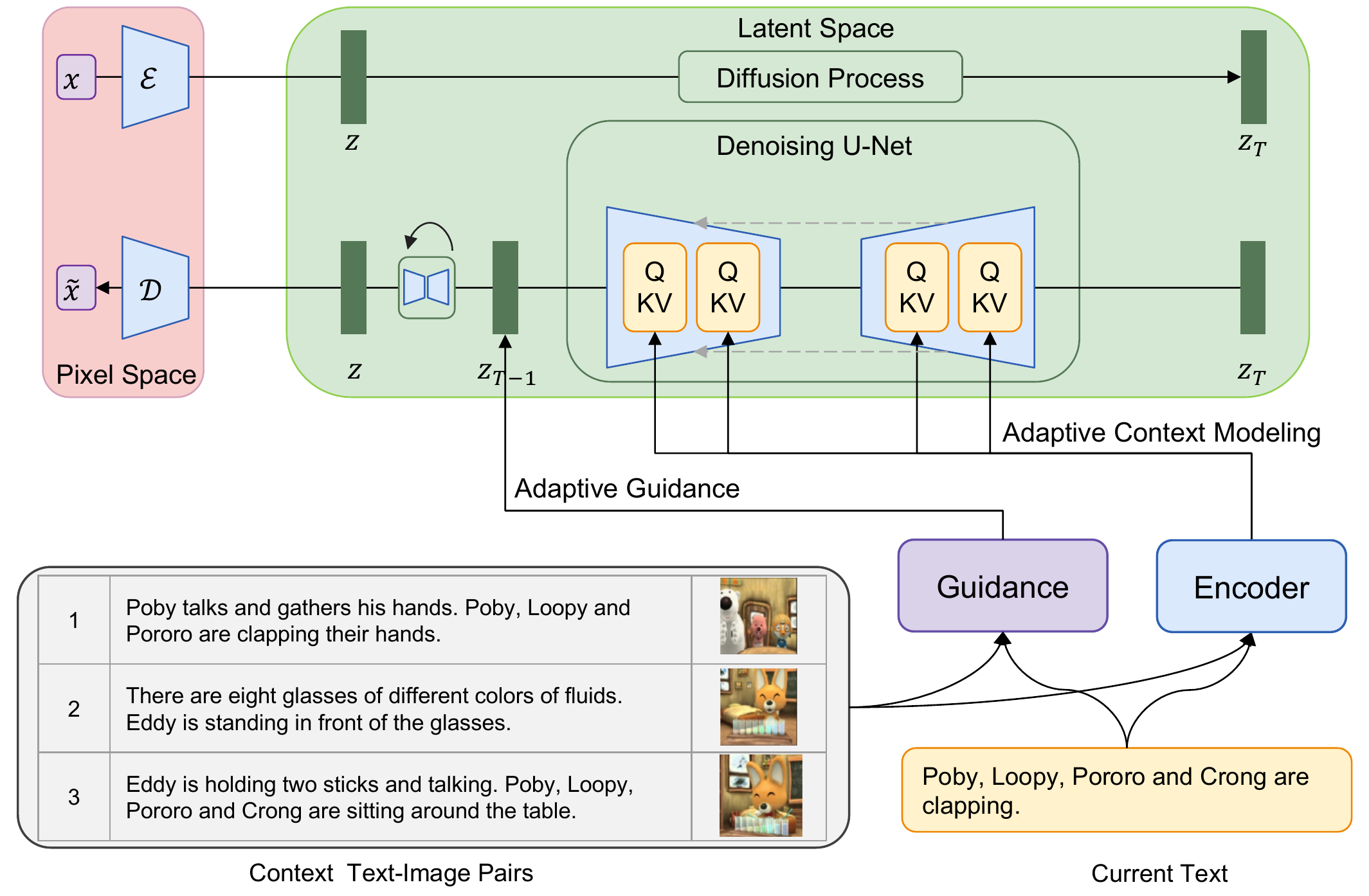}
  \caption{An overview of our model architecture. Based on the latent diffusion model \cite{latentdiffusion}, we propose an adaptive encoder and  an adaptive guidance.
  Adaptive encoder is used to get the adaptive context vectors. Conditional diffusion module transforms context vectors to image. Adaptive guidance aims to guide diffusion sampling process with adaptive context information.}
  \label{fig:model}
\end{figure*}
Unlike the story visualization task, whose input only contains the text story, the story continuation task also includes the first image as input.
\citet{maharana2022storydall} introduces story continuation and modifies the pre-trained text-to-image model DALL-E \cite{ramesh2021zero} by adding a cross attention module for story continuation.
\citet{pan2022synthesizing} employs a history-aware encoding to incorporate previously generated text-image history to diffusion model for visual story generation.

\begin{figure*}[h]
  \includegraphics[width=\linewidth]{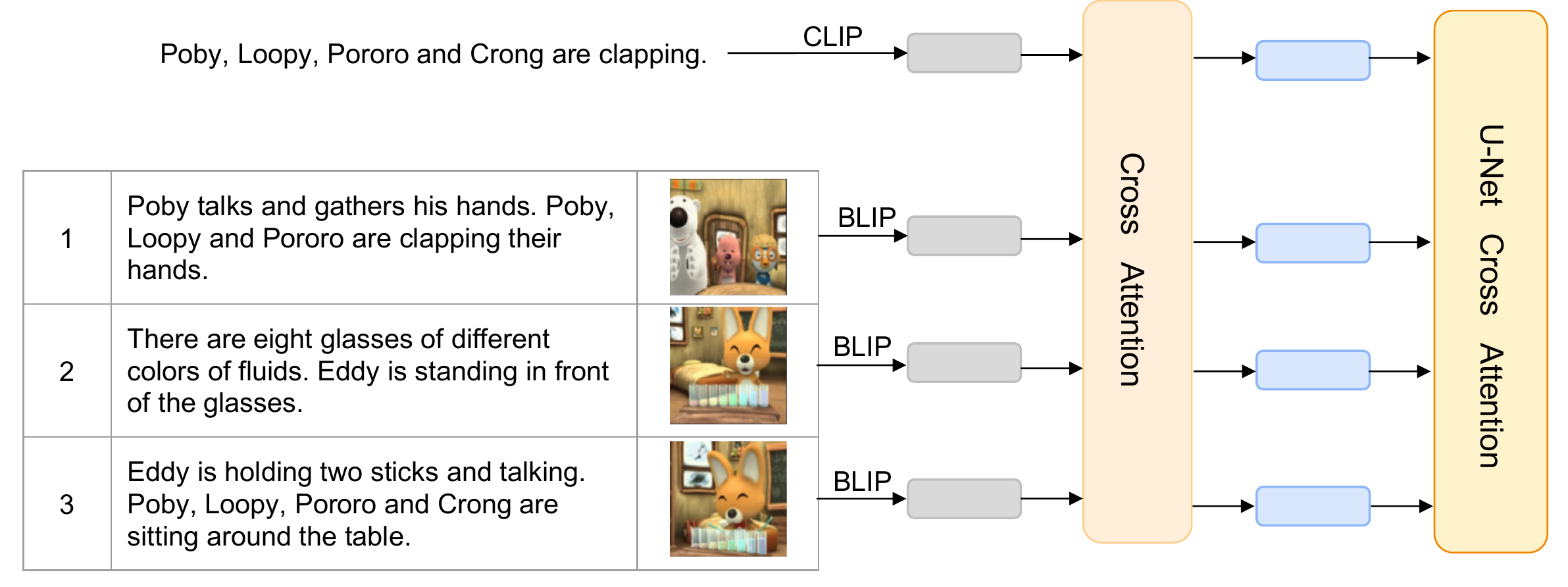}
  \caption{Adaptive encoder consists of three modules: (1) CLIP text encoder is used to encode text prompt. (2) BLIP encoder is used to encode historical text-image pair. (3) Cross attention module is used to filter useful information adaptively.}
  \label{fig:encoder}
\end{figure*}

\section{Model}
We introduce our approach in this section.
We first present the model architecture of our approach (\S \ref{sec:arch}), and then  describe three important components: adaptive encoder (\S \ref{sec:encoder}), conditional diffusion model (\S \ref{sec:diffusion}) and adaptive guidance (\S \ref{sec:guidance}).

\subsection{Model Architecture} \label{sec:arch}
An overview of the approach is depicted in Figure \ref{fig:model}. It includes an adaptive encoder, a conditional diffusion model, and an adaptive guidance.
Based on current text prompt and historical text-image context, the adaptive encoder represents them as conditional vectors.
Then the conditional diffusion model transforms these vectors into the corresponding image. 
During the diffusion sampling process, the adaptive guidance component further guides each diffusion step by comparing it to similar preceding images in the current story to enhance the global consistency of the generated images.

\subsection{Adaptive Encoder} \label{sec:encoder}
Given a story $\mathbf{S}$ which consists of a sequence of text prompts: $\mathbf{S} = \{s_1, s_2, ..., s_L \}$.
Story visualization aims to generate a sequence of images
$\mathbf{X} = \{x_1, x_2, ..., x_L \}$.
Each image corresponds to a text prompt.
Different from text-to-image generation, which only generates one isolated image for the text prompt, story visualization requires global consistency between the generated images.
A natural idea is to combine historical text-image context when generating the current image.

\begin{equation}
\begin{aligned}
    P(\mathbf{X}|\mathbf{S}) &= \prod_{i=1}^L P(x_i \vert \hat{x}_{<i}, \mathbf{S})\\
    &= \prod_{i=1}^L P(x_{i} \vert \tau_{\theta}(\hat{x}_{<i}, s_{\leq i})) \nonumber
\end{aligned}
\end{equation}
where $\tau_{\theta}$ denotes the history-aware conditioning encoder.

As shown in Figure \ref{fig:intro}, we find that some images in the history of the same story are similar to the current image, and some images are even completely irrelevant.
The purpose of the adaptive encoder is to automatically find the relevant historical text-image pairs, and then encode them into the condition vectors.
As shown in Figure \ref{fig:encoder}, adaptive encoder consists of a CLIP text encoder, a BLIP text-image encoder and a cross attention module.
Both CLIP \cite{radford2021learning} and BLIP \cite{li2022blip} are multimodal pre-trained models.
The difference is that CLIP encodes text and image respectively, and BLIP can jointly represent text-image pair.
We use CLIP to get the current text prompt vector $v_i$, and BLIP to get the historical vectors $\{h_0, ..., h_{i-1} \}$.
Then a cross attention is equipped to filter history information and we can obtain the updated vectors $\{\hat{h}_0, ..., h_{i-1} \}$. 
In the cross attention module, the text vector $v_i$ is the query, and each historical vector $h_{<i}$ is the key and value.
Finally, we concatenate the current text vector and history vectors to get the final condition vector
$c = [v_i; \hat{h}_0; ...; \hat{h}_{i-1}]$.

\subsection{Conditional Diffusion Model}\label{sec:diffusion}
Denoising diffusion probabilistic models are a class of score-based generative models, which have recently gained traction in the field of text-to-image generation \cite{ho2020denoising}.
A diffusion model typically contains forward and reverse processes.
Given a data $x_0$ sampled from a real-world data distribution $q(x)$, the forward process is implemented as a predefined Markov chain that gradually corrupts $x_0$ into an isotropic Gaussian distribution $x_T \sim \mathcal{N}(0,I)$ in $T$ steps:
\begin{equation}
    x_t = \sqrt{\alpha_t} x_{t-1} + \sqrt{1-\alpha_t} \epsilon_t, \quad t \in \{1,\dots,T\} \nonumber
\end{equation}
where $\epsilon_t \sim \mathcal{N}(0,I)$, and $\{ \alpha_t \in (0, 1) \}^{T}_{t=1}$ is a predefined noise variance schedule.
The reverse process aims to learn a denoising network $\epsilon_\theta(\cdot)$ to reconstruct the data distribution $x_0$ from the Gaussian noise $x_T \sim \mathcal{N}(0,I)$.
We can express an arbitrary sample $x_t$ from the initial data $x_0$:
\begin{equation}
    x_t = \sqrt{\bar{\alpha}_t} x_0 + \sqrt{1-\bar{\alpha}_t} \epsilon, \nonumber
\end{equation}
where $\bar{\alpha}_t = \prod_{i=1}^t\alpha_i$ and $\epsilon \sim \mathcal{N}(0,I)$. 
The denoising network $\epsilon_\theta(\cdot)$ is trained to recover $x_0$ by predicting the noise $ \epsilon_\theta(x_t, t) $. 
The corresponding learning objective can be formalized as a simple mean-squared error loss between the true noise and the predicted noise:
\begin{equation}
    \mathcal{L} = \mathbb{E}_{x_0, \epsilon , t, c} 
    \left[ || \epsilon - \epsilon_\theta(x_t, t, c) ||_2^2 \right], \nonumber
\end{equation}
where $t$ is uniformly sampled from $\{1, . . . , T\}$, $c$ is condition and $\epsilon \sim \mathcal{N}(0,I)$.

The denoising network $\epsilon_\theta(\cdot)$  is typically implemented by U-Net \cite{ho2020denoising}. To make the diffusion process conditional on the input,
condition $c$ is fed into $\epsilon_\theta(\cdot)$ via a cross-attention layer implementing $Attention(Q, K, V) = softmax(\frac{QK^T}{\sqrt{d}} ) \cdot V$, where the intermediate representations of the U-Net acting as the query $Q$, and the condition embeddings $c$ acting as the key $K$ and value $V$ .

Classifier-free guidance \cite{ho2022classifier} is a widely used technique to improve sample quality while reducing diversity in conditional diffusion models, which jointly trains a single diffusion model on conditional and unconditional objectives via randomly dropping $c$ during training (e.g. with 10\% probability). 
During sampling, the output of the model is extrapolated further in the direction of $\epsilon_{\theta} (x_t|c)$ and away from $\epsilon_{\theta}(x_t|\emptyset)$ as follows:
\begin{equation}
    \hat{\epsilon}_{\theta} (x_t | c) = \epsilon_{\theta}(x_t|\emptyset) + \gamma  \cdot (\epsilon_{\theta} (x_t|c) - \epsilon_{\theta}(x_t|\emptyset)) \nonumber
\end{equation}
where $\gamma \ge1$ is the guidance scale. 
\subsection{Adaptive Guidance} \label{sec:guidance}
Previous work \cite{dhariwal2021diffusion, nichol2021glide, li2022upainting} in text-to-image generation have explored to utilize a classifier or a CLIP \cite{radford2021learning} model to improve a diffusion generator.
A CLIP model consists of two separate pieces: an image encoder and a caption encoder.
The model optimizes a contrastive cross-entropy loss that encourages a high dot-product if the image is paired with the given caption, or a low dot-product if the image and caption correspond to different parts in the training data. 
The denoising diffusion process can be perturbed by the gradient of the dot product of the image and caption.

One of the primary challenges of visual story generation is to maintain consistent background and character appearances throughout the story.
In order to achieve this goal, during the diffusion sample stage, we propose an adaptive guidance, which explicitly requires that the image generated later should be consistent with the preceding generated images. 
Considering that images whose corresponding sentences are similar should have similar scenes/characters.
When generating the image $x_i$ in the story, we first use clip text encoder to calculate the similarity score for each historical text  in $ \{ s_1, ..., s_{i-1}\}$ with the current text $s_i$.
After that, we select the text-image pair with the highest similarity score.
When the similarity score exceeds the threshold, we believe that the selected image and the image to be generated currently have high similarity, and we can use this image to guide the sampling process of the diffusion model.
When the similarity score is lower than the threshold, we think that images in history are not similar to the image to be generated at present, and do not add sampling guidance.

The previous CLIP guided model \cite{nichol2021glide} needs to train an additional noisy CLIP model. 
It's time and computation costly, and difficult to classify noisied image.
Following UPainting \cite{li2022upainting}, we use normal CLIP for guidance, and modify the CLIP inputs as follows:
\begin{gather}
\hat{x}_0 = \frac{1}{\sqrt{\bar{\alpha}_t}} (x_t - \sqrt{1 - \bar{\alpha}_t} \epsilon_t) \nonumber \\
x_{in} = \sqrt{1 - \bar{\alpha}_t} \hat{x}_0 + (1 - \sqrt{1 - \bar{\alpha}_t}) x_t \nonumber
\end{gather}
The denoising diffusion process can be formulated as follows:
\begin{equation}
\begin{split}
        \hat{\epsilon}_{\theta}' (x_t | c) = \epsilon_{\theta}(x_t|\emptyset) + \gamma  \cdot (\epsilon_{\theta} (x_t|c) - \epsilon_{\theta}(x_t|\emptyset)) \\
        - g \sqrt{1 - \bar{\alpha}_{t}} \nabla_{x_t} (f(x_{in}) \cdot f(x_h)) \nonumber
\end{split}
\label{eq_combine}
\end{equation}
\noindent where $\gamma \ge 1$ is the classifier weight, $g \ge 0$ is adaptive guidance weight, $f(.)$ is the CLIP image encoder and $x_h$ is the most similar image to the current image $x_t$.

\section{Experiments}
\subsection{Datasets and Metrics}
\begin{figure}[!t]
  \includegraphics[width=\linewidth]{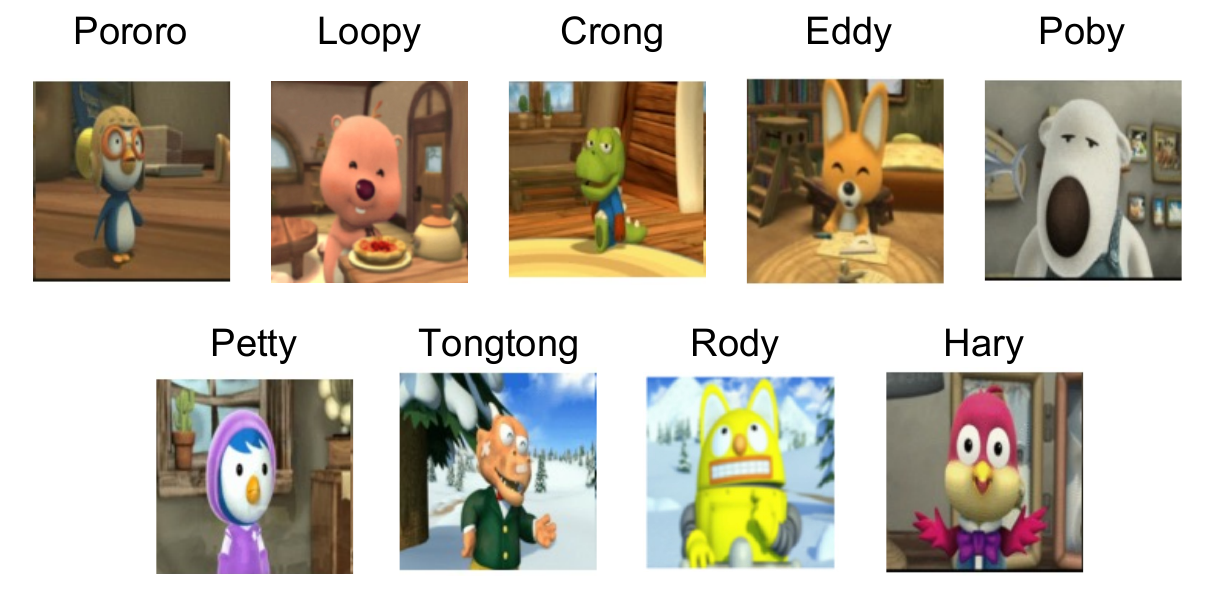}
  \caption{Main characters in PororoSV dataset.}
  \label{fig:pororo_char}
\end{figure}

\begin{figure}[!t]
  \includegraphics[width=\linewidth]{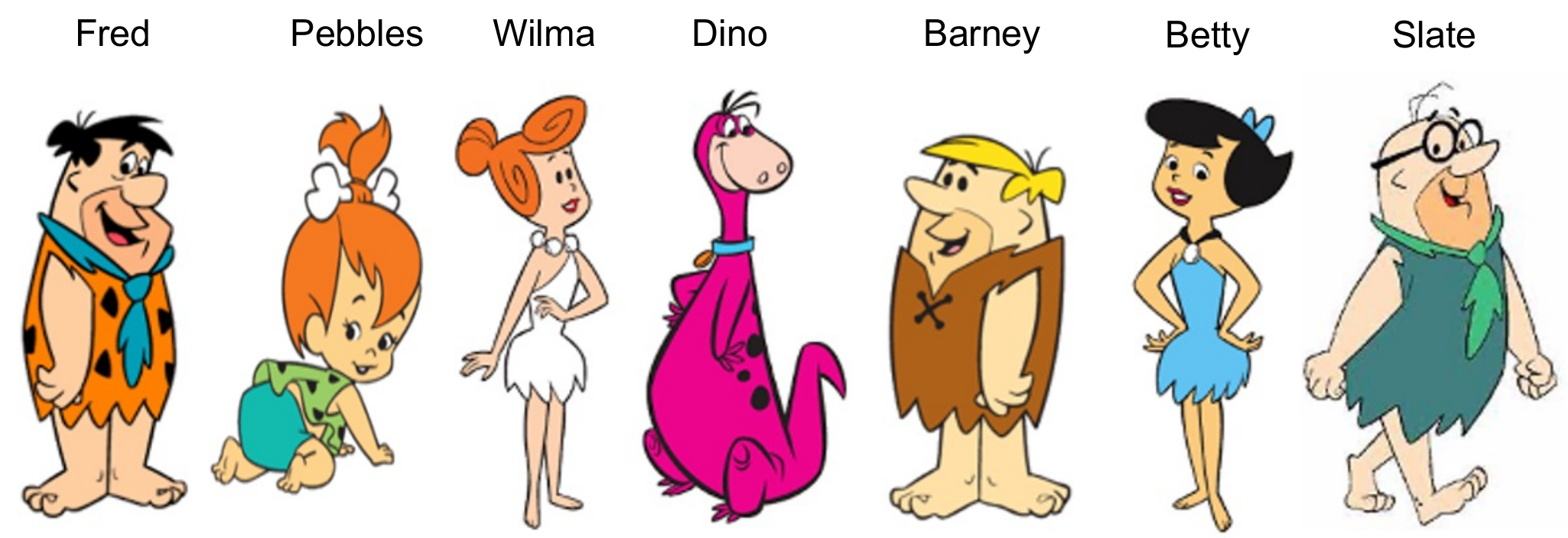}
  \caption{Main characters in FlintstonesSV dataset.}
  \label{fig:flint_char}
\end{figure}

We carried out experiments on both story visualization and story continuation tasks.
Given a sequence of sentences forming a narrative, story visualization is the task of generating a corresponding sequence of images.
Story continuation, including an initial ground truth image as input, is a variant of story visualization.
We use two popular datasets, PororoSV \cite{li2019storygan} and FlintstonesSV \cite{TanmayGupta2018ImagineTS}, to evaluate our model.
We give the statistics of two datasets in Table \ref{tab:datasets} and show  main characters in Figure \ref{fig:pororo_char}  and Figure \ref{fig:flint_char} to help the reader understand our examples.
\begin{table}[h]
\centering
\begin{tabular}{lccc}
\toprule
\bf & Train & Valid & Test \\
\midrule
PororoSV & 10,191	& 2,334	&2,208 \\
FlintstonesSV & 20,132 & 2,071	& 2,309\\
\bottomrule
\end{tabular}
\caption{Statistics for PororoSV and FlintstonesSV datasets.}
\label{tab:datasets}
\end{table}

We adopt the automatic evaluation metrics following existing works and report results using the evaluation script provided in prior work\footnote{https://github.com/adymaharana/VLCStoryGan}.
\paragraph{Frechet Inception Distance (FID)} captures the level of similarity between two groups based on statistical analysis of visual features in their respective raw images, using the inception v3 model. Lower FID scores indicate higher resemblance between the predicted images and the ground-truth images.
\paragraph{Character F1 score}  calculates the proportion of characters present in the generated images that exactly match the characters in the story inputs. To achieve this, a pretrained inception v3 model \cite{ChristianSzegedy2015RethinkingTI} is fine-tuned on each dataset using a multi-label classification loss, enabling it to make predictions of characters in test images.
\paragraph{Frame Accuracy} evaluates whether all characters from a story are correctly represented in the corresponding images, utilizing the same model employed in the Character F1 score. While the Character F1 score measures the proportion of characters captured in a story, Frame Accuracy quantifies the percentage of samples where all characters are appropriately included.

\subsection{Implementation Details}
Our model is fine-tuned from the pre-traiend Stable Diffusion text-to-image generation model.
We use CLIP base model and BLIP base model.
We only train the parameters of diffusion model and cross attention module, and freeze the parameters of variational auto-encoder, CLIP and BLIP, which could speed up training and save GPU memory.
Follow previous work, we train our model for 50 epochs. 
We use Adam optimizer and set learning rate to 1e-4.
For $\gamma$, we used the default value in stable diffusion without adjustment. For the threshold of similarity score, we randomly sampled 50 stories to calculate the similarity score, then manually observed the relationship between the similarity score and the image similarity, and finally set the threshold to 0.65. For the $g$ , we chose the best value 0.15 from {0.1, 0.15, 0.2, 0.5}.

\subsection{Baselines}
\paragraph{StoryGAN} \cite{li2019storygan} uses the standard GAN technique, which includes a recurrent text encoder, an image generation module, and two discriminators - image and story discriminator.

\paragraph{StoryGANc} \cite{maharana2022storydall} follows the general framework of the StoryGAN model and adds the source image as input for the story continuation task.

\paragraph{CP-CSV} \cite{YunZhuSong2020CharacterPreservingCS} tries to better preserve character information with three modules: story and context encoder, figure-ground segmentation, and figure-ground aware generation.

\paragraph{DUCO-StoryGAN} \cite{AdyashaMaharana2021ImprovingGA} utlizes a video captioning model to generate an additional learning signal forcing the alignment of image and text, and a memory-augmented transformer to model complex interactions between frames.

\paragraph{VLC-StoryGAN} \cite{AdyashaMaharana2021IntegratingVL} incorporates constituency parse trees, commonsense information and visual information, including bounding boxes and dense captioning, to enhance the visual quality and image consistency.

\paragraph{Word-Level} \cite{BowenLi2022WordLevelFS} incorporates word information and extends word-level spatial attention to focus on all words and visual spatial locations in the entire story.

\paragraph{StoryDALL-E} \cite{maharana2022storydall} modifies the pre-trained text-to-image model DALL-E by adding a cross attention module for story continuation.

\paragraph{AR-LDM} \cite{pan2022synthesizing} employs a history-aware encoding module to incorporate the current text prompt and previously generated text-image history  to diffusion model for visual story generation.

\begin{table*}[ht]
\centering
\begin{tabular}{lcccccc}
\toprule
\multirow{2}{*}{\textbf{Model}} & \multicolumn{3}{c}{\textbf{ PororoSV }} & \multicolumn{3}{c}{\textbf{ FlintstonesSV }}  \\
\cline{2-7}
 & FID $\downarrow$ &{Char-F1$\uparrow$}& F-Acc$\uparrow$& FID $\downarrow$&{Char-F1$\uparrow$}& F-Acc$\uparrow$\\
\midrule
 StoryGANc(BERT) & 72.98 & 43.22 & 17.09 & 91.37 & 70.45 & 55.78  \\
 StoryGANc (CLIP) & 74.63 & 39.68 & 16.57 & 90.29 & 72.80 & 58.39 \\
 StoryDALL-E(prompt) & 61.23 & 29.68 & 11.65 & 53.71 & 42.48 & 32.54  \\
StoryDALL-E (finetuning) & 25.90 & 36.97 & 17.26 & 26.49 & 73.43 & 55.19  \\
MEGA-StoryDALL-E & 23.48 & 39.91 &  18.01 &  23.58 &  74.26 &  54.68 \\
AR-LDM & 17.40 & - & - & 19.28 & - & - \\
\textbf{ACM-VSG (Ours)}  & \textbf{15.36} & \textbf{45.71} & \textbf{22.62} &\textbf{18.41} & \textbf{94.95} & \textbf{88.89} \\
\bottomrule
\end{tabular}
\caption{\label{tab:test_fid}Results on the test sets of PororoSV and FlintstonesSV datasets from various models. Scores are based on FID , character classification F1, and frame accuracy evaluations.}
\label{tab:results_sc}
\end{table*}

\begin{table}[h]
\centering
\begin{tabular}{p{5cm} c}
\toprule
\bf Model & FID $\downarrow$ \\
\midrule
StoryGAN &  158.06 \\
CP-CSV  & 149.29  \\
DUCO-StoryGAN  &  96.51 \\
VLC-StoryGAN  & 84.96 \\
VP-CSV  & 65.51 \\
Word-Level SV  & 56.08 \\
AR-LDM  &  16.59 \\
\textbf{ACM-VSG (Ours)} & \textbf{15.48} \\
\bottomrule
\end{tabular}
\caption{Story visualization FID score results on PororoSV dataset.}
\label{tab:results_sv}
\end{table}

\begin{figure*}[!h]
  \includegraphics[width=\linewidth]{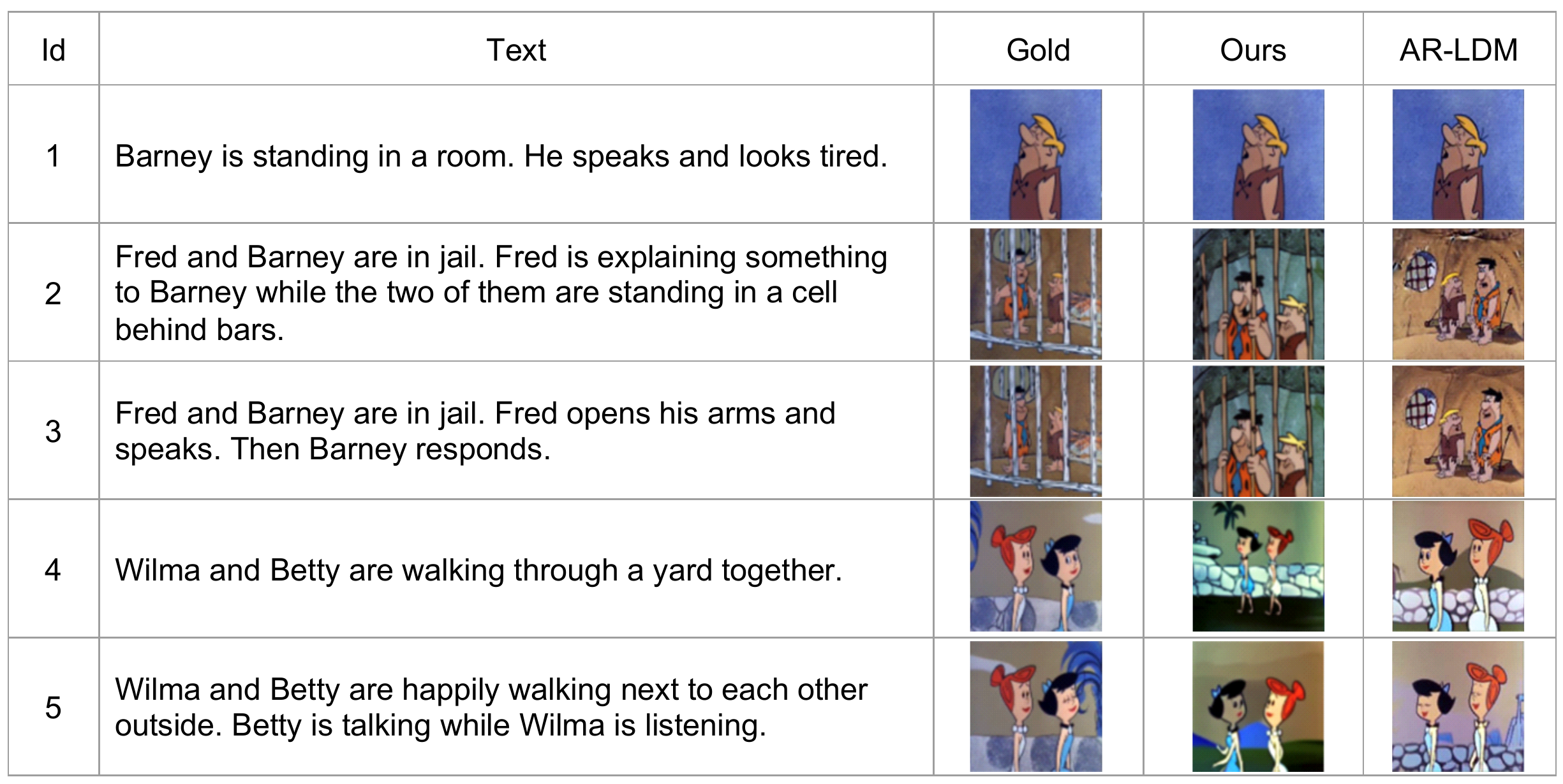}
  \caption{Example of generated images from previous model AR-LDM and our proposed model.}
  \label{fig:compare_case}
\end{figure*}
\section{Results}

\subsection{Story Visualization}
We evaluate our model on PororoSV dataset for story visualization task. Results are shown in Table \ref{tab:results_sv}.
We can observe that diffusion-based model outperforms the prior methods by a large margin, and our proposed ACM-VSG achieves the best FID score 15.48, indicating our model is able to generate high-quality images.

\begin{figure*}[!h]
  \includegraphics[width=\linewidth]{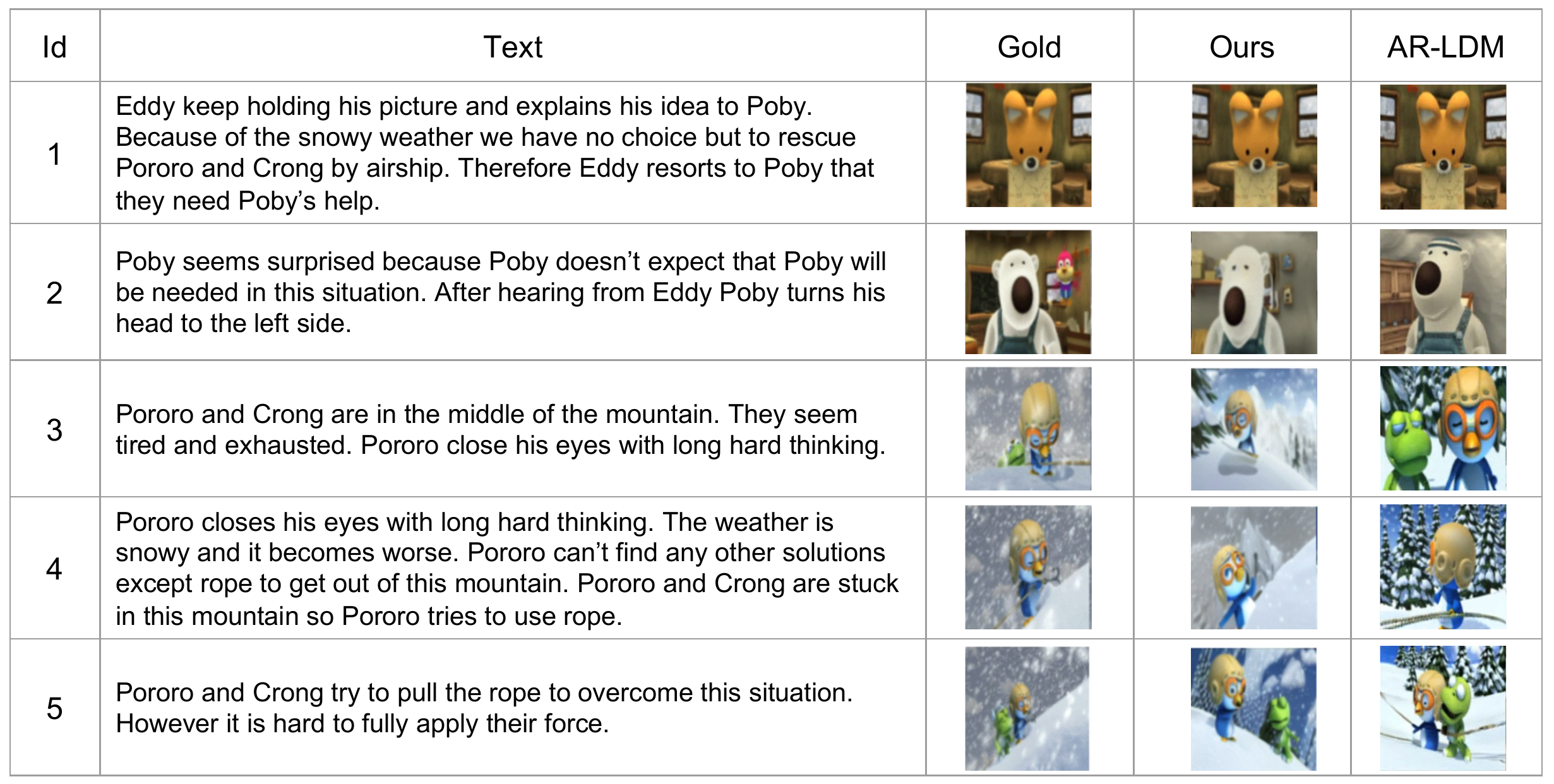}
  \caption{Example of generated images from previous model AR-LDM and our proposed model.}
  \label{fig:compare_case2}
\end{figure*}

\subsection{Story Continuation}
Table \ref{tab:results_sc} shows the results for story continuation task.
As we can see, our model can achieve the best results on both datasets,  15.36 and 18.41 FID for PororoSV and FlintstonesSV, respectively.
And our model can greatly preserve characters to improve the consistency of the story.
In addition, we show an example on FlintstonesSV and pororoSV dataset in Figure \ref{fig:compare_case} and Figure \ref{fig:compare_case2}. 
We can observe that our model is able to maintain the text-image alignment and consistency across images.

\subsection{Ablation Study}

Table \ref{tab:ablation } shows ablation studies to ensure that each component in the our proposed method benefits visual story  generation. \text{-Guidance} means removing the adaptive guidance.
\text{-Attention} means removing the cross attention module in the adaptive encoder.

\begin{table}[t]
\centering
\begin{tabular}{lccc}
\toprule
\bf Model & FID $\downarrow$ & Char-F1$\uparrow$ & F-ACC$\uparrow$  \\
\midrule
ACM-VSG & 15.36 & 45.71 & 22.62 \\
\quad - Guidance & 15.96 & 44.56 & 22.13 \\
\quad - Attention & 16.88 & 44.27  & 20.25 \\
\bottomrule
\end{tabular}
\caption{Ablation study results for story continuation task on PororoSV.}
\label{tab:ablation }
\end{table}

\begin{figure}[!h]
  \includegraphics[width=\linewidth]{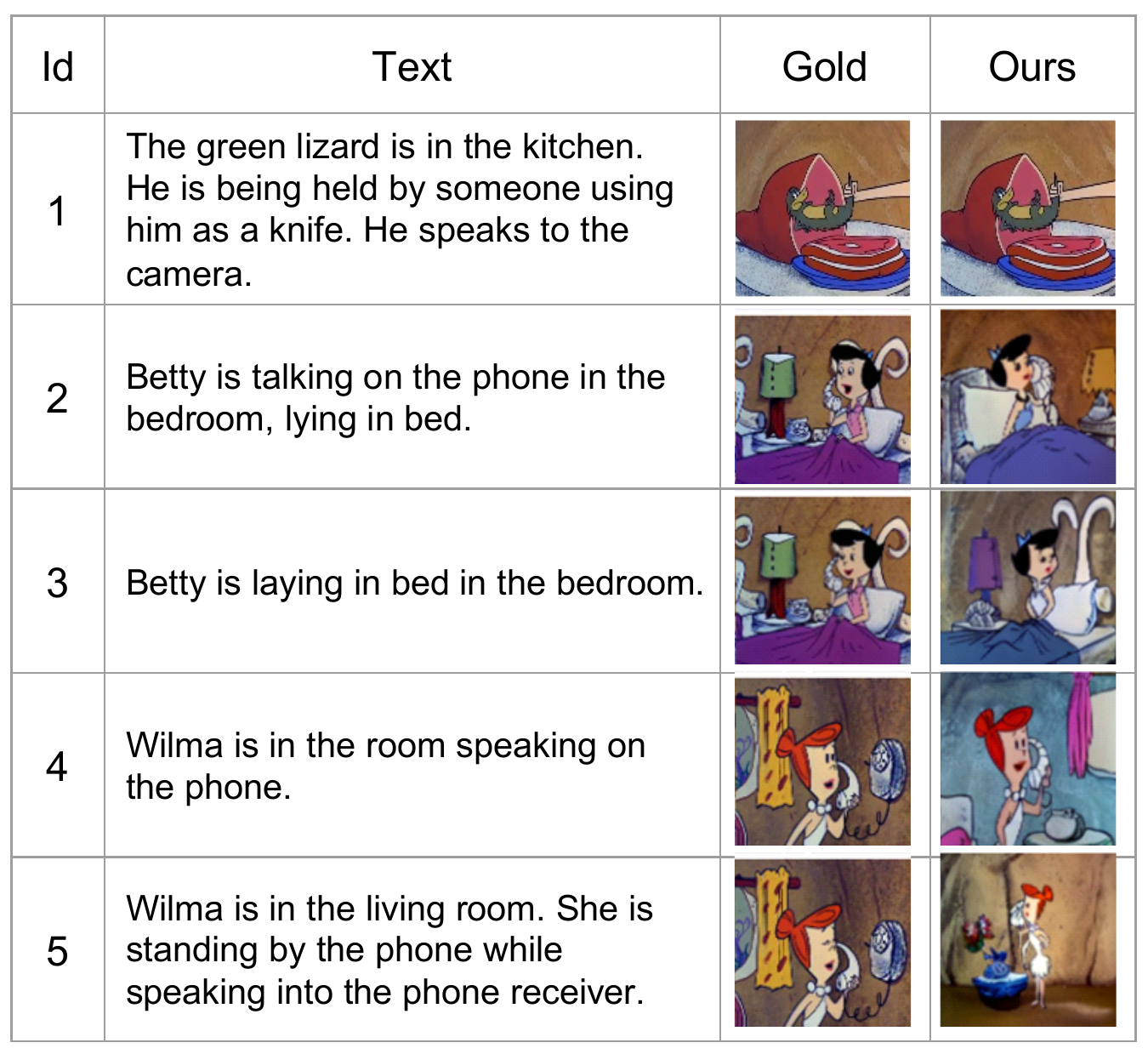}
  \caption{The generated image story has global inconsistency.}
  \label{fig:bad_case1}
\end{figure}

\begin{figure}[!h]
  \centering
  \includegraphics[width=\linewidth]{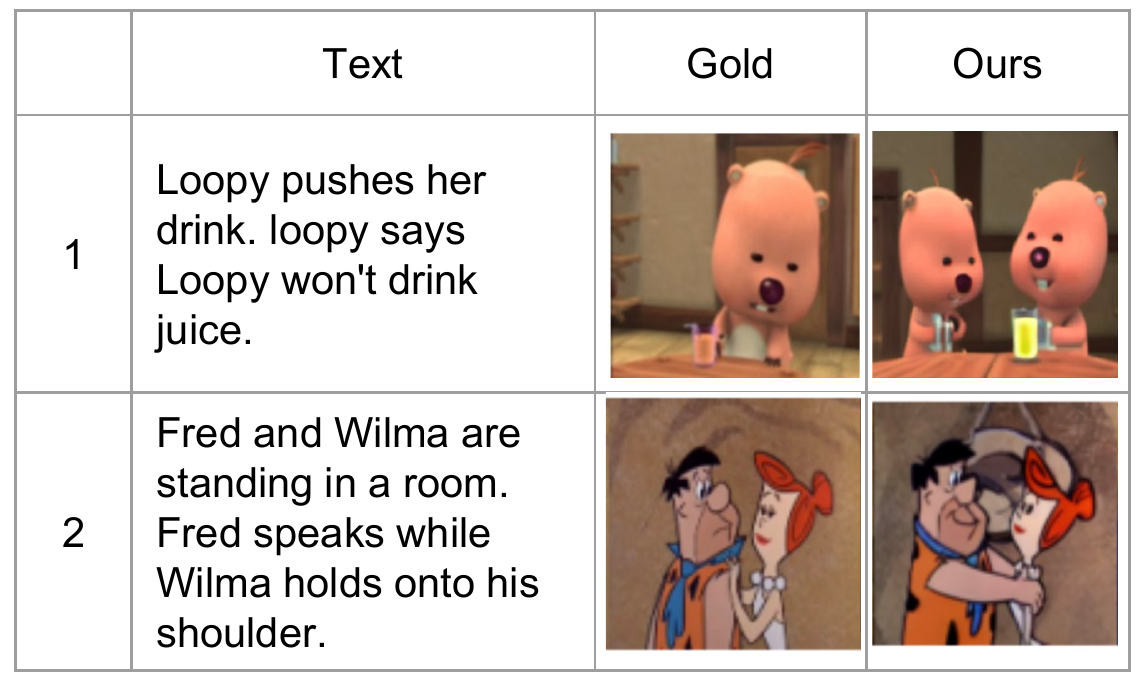}
  \caption{The bad cases for repetitive character and character action error.}
  \label{fig:bad_case23}
\end{figure}

\subsection{Error Analysis}
Our model significantly improves the performance of visual story generation, but there are still some limits.
In order to solve these limitations in future work, we analyze the generated images.
We randomly sample hundreds of stories from PororoSV and FlintstonesSV datasets and summarize the errors.

\paragraph{Story Inconsistency.}
As shown in Figure \ref{fig:bad_case1}, there are inconsistencies across generated images.
The style of the bed and the color of the quilt between the second and third generated images are inconsistent.
The environment around \textit{Wilma} is inconsistent between the fourth and fifth images.

\paragraph{Repetitive Character.}
As shown in Figure \ref{fig:bad_case23} case 1, the model may generate the same character repeatedly if it appears multiple times in the text.

\paragraph{Character Action Error.}
As shown in Figure \ref{fig:bad_case23} case 2, the subtle actions of  characters in the image are misaligned with the text.
In case 2, Wilma holds onto Fred's shoulder in text, while Fred holds onto Wilma's shoulder in the generated image.

\section{Conclusion}
In this paper, we explore an effective adaptive context modeling method to improve visual story generation.
First, we use an adaptive encoder to select the context closely related to the current image from the historical text-image pairs using the current text.
Then we fed the context vectors and the text vector to the diffusion model,
and use an adaptive guidance to guide the generation of the current image.
Experimental results verify that adaptive context modeling could help generate higher quality images and more consistent stories.
In addition, we analyze the generated images and find potential research directions in the future: (1) focus on the global consistency of the story, (2) pay attention to the action and expression of the character, (3) obtain the exact semantics of long stories.

\section*{Limitations}
A limitation of this work is that it is only evaluated on synthesized datasets of cartoons with limited characters and scenes. 
In the real world application, there might be many different scenes/characters, posing new challenges to the proposed approach. 
Another limitation is the requirement of supervised training data and resources. 
Despite the number of trainable parameters of our approach (850M) is less than AR-LDM ($\sim$1.5B), the model still needs many story-level training data and computing resources. 
\section*{Acknowledgements}
Xiaocheng Feng is the corresponding author of this work. We thank the anonymous reviewers for their insightful comments. 
Zhangyin Feng, Xinmiao Yu, Xiaocheng Feng and Bing Qin are supported by the National Key R\&D Program of China via grant 2020AAA0106502, National Natural Science Foundation of China (NSFC) via grant 62276078, the Key R\&D Program of Heilongjiang via grant 2022ZX01A32 and the International Cooperation Project of PCL, PCL2022D01.

\bibliography{anthology,custom}

\begin{thebibliography}{37}
\expandafter\ifx\csname natexlab\endcsname\relax\def\natexlab#1{#1}\fi

\bibitem[{Balaji et~al.(2022)Balaji, Nah, Huang, Vahdat, Song, Kreis, Aittala,
  Aila, Laine, Catanzaro et~al.}]{balaji2022ediffi}
Yogesh Balaji, Seungjun Nah, Xun Huang, Arash Vahdat, Jiaming Song, Karsten
  Kreis, Miika Aittala, Timo Aila, Samuli Laine, Bryan Catanzaro, et~al. 2022.
\newblock ediffi: Text-to-image diffusion models with an ensemble of expert
  denoisers.
\newblock \emph{arXiv preprint arXiv:2211.01324}.

\bibitem[{Bao et~al.(2022)Bao, Li, Cao, and Zhu}]{bao2022all}
Fan Bao, Chongxuan Li, Yue Cao, and Jun Zhu. 2022.
\newblock All are worth words: a vit backbone for score-based diffusion models.
\newblock \emph{arXiv preprint arXiv:2209.12152}.

\bibitem[{Chang et~al.(2023)Chang, Zhang, Barber, Maschinot, Lezama, Jiang,
  Yang, Murphy, Freeman, Rubinstein et~al.}]{chang2023muse}
Huiwen Chang, Han Zhang, Jarred Barber, AJ~Maschinot, Jose Lezama, Lu~Jiang,
  Ming-Hsuan Yang, Kevin Murphy, William~T Freeman, Michael Rubinstein, et~al.
  2023.
\newblock Muse: Text-to-image generation via masked generative transformers.
\newblock \emph{arXiv preprint arXiv:2301.00704}.

\bibitem[{Dhariwal and Nichol(2021)}]{dhariwal2021diffusion}
Prafulla Dhariwal and Alexander Nichol. 2021.
\newblock Diffusion models beat gans on image synthesis.
\newblock \emph{Advances in Neural Information Processing Systems},
  34:8780--8794.

\bibitem[{Ding et~al.(2021)Ding, Yang, Hong, Zheng, Zhou, Yin, Lin, Zou, Shao,
  Yang et~al.}]{ding2021cogview}
Ming Ding, Zhuoyi Yang, Wenyi Hong, Wendi Zheng, Chang Zhou, Da~Yin, Junyang
  Lin, Xu~Zou, Zhou Shao, Hongxia Yang, et~al. 2021.
\newblock Cogview: Mastering text-to-image generation via transformers.
\newblock \emph{Advances in Neural Information Processing Systems},
  34:19822--19835.

\bibitem[{Feng et~al.(2022)Feng, Zhang, Yu, Fang, Li, Chen, Lu, Liu, Yin, Feng
  et~al.}]{feng2022ernie}
Zhida Feng, Zhenyu Zhang, Xintong Yu, Yewei Fang, Lanxin Li, Xuyi Chen, Yuxiang
  Lu, Jiaxiang Liu, Weichong Yin, Shikun Feng, et~al. 2022.
\newblock Ernie-vilg 2.0: Improving text-to-image diffusion model with
  knowledge-enhanced mixture-of-denoising-experts.
\newblock \emph{arXiv preprint arXiv:2210.15257}.

\bibitem[{Gafni et~al.(2022)Gafni, Polyak, Ashual, Sheynin, Parikh, and
  Taigman}]{gafni2022make}
Oran Gafni, Adam Polyak, Oron Ashual, Shelly Sheynin, Devi Parikh, and Yaniv
  Taigman. 2022.
\newblock Make-a-scene: Scene-based text-to-image generation with human priors.
\newblock \emph{arXiv preprint arXiv:2203.13131}.

\bibitem[{Goodfellow et~al.(2014)Goodfellow, Pouget-Abadie, Mirza, Xu,
  Warde-Farley, Ozair, Courville, and Bengio}]{goodfellow2014generative}
Ian Goodfellow, Jean Pouget-Abadie, Mehdi Mirza, Bing Xu, David Warde-Farley,
  Sherjil Ozair, Aaron Courville, and Yoshua Bengio. 2014.
\newblock Generative adversarial nets.
\newblock \emph{Advances in neural information processing systems}, 27.

\bibitem[{Gupta et~al.(2018)Gupta, Schwenk, Farhadi, Hoiem, and
  Kembhavi}]{TanmayGupta2018ImagineTS}
Tanmay Gupta, Dustin Schwenk, Ali Farhadi, Derek Hoiem, and Aniruddha Kembhavi.
  2018.
\newblock Imagine this! scripts to compositions to videos.
\newblock \emph{arXiv: Computer Vision and Pattern Recognition}.

\bibitem[{Ho et~al.(2020)Ho, Jain, and Abbeel}]{ho2020denoising}
Jonathan Ho, Ajay Jain, and Pieter Abbeel. 2020.
\newblock Denoising diffusion probabilistic models.
\newblock \emph{Advances in Neural Information Processing Systems},
  33:6840--6851.

\bibitem[{Ho and Salimans(2022)}]{ho2022classifier}
Jonathan Ho and Tim Salimans. 2022.
\newblock Classifier-free diffusion guidance.
\newblock \emph{arXiv preprint arXiv:2207.12598}.

\bibitem[{Li and Lukasiewicz(2022)}]{BowenLi2022WordLevelFS}
Bowen Li and Thomas Lukasiewicz. 2022.
\newblock Word-level fine-grained story visualization.

\bibitem[{Li et~al.(2020)Li, Kong, and Zhou}]{li2020improved}
Chunye Li, Liya Kong, and Zhiping Zhou. 2020.
\newblock Improved-storygan for sequential images visualization.
\newblock \emph{Journal of Visual Communication and Image Representation},
  73:102956.

\bibitem[{Li et~al.(2022{\natexlab{a}})Li, Li, Xiong, and Hoi}]{li2022blip}
Junnan Li, Dongxu Li, Caiming Xiong, and Steven Hoi. 2022{\natexlab{a}}.
\newblock Blip: Bootstrapping language-image pre-training for unified
  vision-language understanding and generation.
\newblock \emph{arXiv preprint arXiv:2201.12086}.

\bibitem[{Li et~al.(2022{\natexlab{b}})Li, Xu, Xiao, Liu, Yang, Li, Wang, Feng,
  She, Lyu et~al.}]{li2022upainting}
Wei Li, Xue Xu, Xinyan Xiao, Jiachen Liu, Hu~Yang, Guohao Li, Zhanpeng Wang,
  Zhifan Feng, Qiaoqiao She, Yajuan Lyu, et~al. 2022{\natexlab{b}}.
\newblock Upainting: Unified text-to-image diffusion generation with
  cross-modal guidance.
\newblock \emph{arXiv preprint arXiv:2210.16031}.

\bibitem[{Li et~al.(2019)Li, Gan, Shen, Liu, Cheng, Wu, Carin, Carlson, and
  Gao}]{li2019storygan}
Yitong Li, Zhe Gan, Yelong Shen, Jingjing Liu, Yu~Cheng, Yuexin Wu, Lawrence
  Carin, David Carlson, and Jianfeng Gao. 2019.
\newblock Storygan: A sequential conditional gan for story visualization.
\newblock In \emph{Proceedings of the IEEE/CVF Conference on Computer Vision
  and Pattern Recognition}, pages 6329--6338.

\bibitem[{Maharana and
  Bansal(2021{\natexlab{a}})}]{maharana-bansal-2021-integrating}
Adyasha Maharana and Mohit Bansal. 2021{\natexlab{a}}.
\newblock \href {https://doi.org/10.18653/v1/2021.emnlp-main.543} {Integrating
  visuospatial, linguistic, and commonsense structure into story
  visualization}.
\newblock In \emph{Proceedings of the 2021 Conference on Empirical Methods in
  Natural Language Processing}, pages 6772--6786, Online and Punta Cana,
  Dominican Republic. Association for Computational Linguistics.

\bibitem[{Maharana and
  Bansal(2021{\natexlab{b}})}]{AdyashaMaharana2021IntegratingVL}
Adyasha Maharana and Mohit Bansal. 2021{\natexlab{b}}.
\newblock Integrating visuospatial, linguistic and commonsense structure into
  story visualization.
\newblock \emph{arXiv: Computation and Language}.

\bibitem[{Maharana et~al.(2021{\natexlab{a}})Maharana, Hannan, and
  Bansal}]{maharana-etal-2021-improving}
Adyasha Maharana, Darryl Hannan, and Mohit Bansal. 2021{\natexlab{a}}.
\newblock \href {https://doi.org/10.18653/v1/2021.naacl-main.194} {Improving
  generation and evaluation of visual stories via semantic consistency}.
\newblock In \emph{Proceedings of the 2021 Conference of the North American
  Chapter of the Association for Computational Linguistics: Human Language
  Technologies}, pages 2427--2442, Online. Association for Computational
  Linguistics.

\bibitem[{Maharana et~al.(2021{\natexlab{b}})Maharana, Hannan, and
  Bansal}]{AdyashaMaharana2021ImprovingGA}
Adyasha Maharana, Darryl Hannan, and Mohit Bansal. 2021{\natexlab{b}}.
\newblock Improving generation and evaluation of visual stories via semantic
  consistency.
\newblock \emph{arXiv: Computation and Language}.

\bibitem[{Maharana et~al.(2022)Maharana, Hannan, and
  Bansal}]{maharana2022storydall}
Adyasha Maharana, Darryl Hannan, and Mohit Bansal. 2022.
\newblock Storydall-e: Adapting pretrained text-to-image transformers for story
  continuation.
\newblock In \emph{European Conference on Computer Vision}, pages 70--87.
  Springer.

\bibitem[{Nichol et~al.(2021)Nichol, Dhariwal, Ramesh, Shyam, Mishkin, McGrew,
  Sutskever, and Chen}]{nichol2021glide}
Alex Nichol, Prafulla Dhariwal, Aditya Ramesh, Pranav Shyam, Pamela Mishkin,
  Bob McGrew, Ilya Sutskever, and Mark Chen. 2021.
\newblock Glide: Towards photorealistic image generation and editing with
  text-guided diffusion models.
\newblock \emph{arXiv preprint arXiv:2112.10741}.

\bibitem[{Pan et~al.(2022)Pan, Qin, Li, Xue, and Chen}]{pan2022synthesizing}
Xichen Pan, Pengda Qin, Yuhong Li, Hui Xue, and Wenhu Chen. 2022.
\newblock Synthesizing coherent story with auto-regressive latent diffusion
  models.
\newblock \emph{arXiv preprint arXiv:2211.10950}.

\bibitem[{Radford et~al.(2021)Radford, Kim, Hallacy, Ramesh, Goh, Agarwal,
  Sastry, Askell, Mishkin, Clark et~al.}]{radford2021learning}
Alec Radford, Jong~Wook Kim, Chris Hallacy, Aditya Ramesh, Gabriel Goh,
  Sandhini Agarwal, Girish Sastry, Amanda Askell, Pamela Mishkin, Jack Clark,
  et~al. 2021.
\newblock Learning transferable visual models from natural language
  supervision.
\newblock In \emph{International Conference on Machine Learning}, pages
  8748--8763. PMLR.

\bibitem[{Ramesh et~al.(2022)Ramesh, Dhariwal, Nichol, Chu, and Chen}]{dalle-2}
Aditya Ramesh, Prafulla Dhariwal, Alex Nichol, Casey Chu, and Mark Chen. 2022.
\newblock Hierarchical text-conditional image generation with clip latents.
\newblock \emph{arXiv preprint arXiv:2204.06125}.

\bibitem[{Ramesh et~al.(2021{\natexlab{a}})Ramesh, Pavlov, Goh, Gray, Voss,
  Radford, Chen, and Sutskever}]{dalle}
Aditya Ramesh, Mikhail Pavlov, Gabriel Goh, Scott Gray, Chelsea Voss, Alec
  Radford, Mark Chen, and Ilya Sutskever. 2021{\natexlab{a}}.
\newblock Zero-shot text-to-image generation.
\newblock In \emph{International Conference on Machine Learning}, pages
  8821--8831. PMLR.

\bibitem[{Ramesh et~al.(2021{\natexlab{b}})Ramesh, Pavlov, Goh, Gray, Voss,
  Radford, Chen, and Sutskever}]{ramesh2021zero}
Aditya Ramesh, Mikhail Pavlov, Gabriel Goh, Scott Gray, Chelsea Voss, Alec
  Radford, Mark Chen, and Ilya Sutskever. 2021{\natexlab{b}}.
\newblock Zero-shot text-to-image generation.
\newblock In \emph{International Conference on Machine Learning}, pages
  8821--8831. PMLR.

\bibitem[{Reed et~al.(2016)Reed, Akata, Yan, Logeswaran, Schiele, and
  Lee}]{reed2016generative}
Scott Reed, Zeynep Akata, Xinchen Yan, Lajanugen Logeswaran, Bernt Schiele, and
  Honglak Lee. 2016.
\newblock Generative adversarial text to image synthesis.
\newblock In \emph{International conference on machine learning}, pages
  1060--1069. PMLR.

\bibitem[{Rombach et~al.(2022)Rombach, Blattmann, Lorenz, Esser, and
  Ommer}]{latentdiffusion}
Robin Rombach, Andreas Blattmann, Dominik Lorenz, Patrick Esser, and Bj{\"o}rn
  Ommer. 2022.
\newblock High-resolution image synthesis with latent diffusion models.
\newblock In \emph{Proceedings of the IEEE/CVF Conference on Computer Vision
  and Pattern Recognition}, pages 10684--10695.

\bibitem[{Saharia et~al.(2022)Saharia, Chan, Saxena, Li, Whang, Denton,
  Ghasemipour, Ayan, Mahdavi, Lopes et~al.}]{imagen}
Chitwan Saharia, William Chan, Saurabh Saxena, Lala Li, Jay Whang, Emily
  Denton, Seyed Kamyar~Seyed Ghasemipour, Burcu~Karagol Ayan, S~Sara Mahdavi,
  Rapha~Gontijo Lopes, et~al. 2022.
\newblock Photorealistic text-to-image diffusion models with deep language
  understanding.
\newblock \emph{arXiv preprint arXiv:2205.11487}.

\bibitem[{Song et~al.(2020)Song, Tam, Chen, Lu, and
  Shuai}]{YunZhuSong2020CharacterPreservingCS}
Yun~Zhu Song, Zhi~Rui Tam, Hung~Jen Chen, Huiao~Han Lu, and Hong-Han Shuai.
  2020.
\newblock Character-preserving coherent story visualization.
\newblock \emph{Springer International Publishing eBooks}.

\bibitem[{Szegedy et~al.(2015)Szegedy, Vanhoucke, Ioffe, Shlens, and
  Wojna}]{ChristianSzegedy2015RethinkingTI}
Christian Szegedy, Vincent Vanhoucke, Sergey Ioffe, Jonathon Shlens, and
  Zbigniew Wojna. 2015.
\newblock Rethinking the inception architecture for computer vision.
\newblock \emph{arXiv: Computer Vision and Pattern Recognition}.

\bibitem[{Van Den~Oord et~al.(2017)Van Den~Oord, Vinyals et~al.}]{van2017dvae}
Aaron Van Den~Oord, Oriol Vinyals, et~al. 2017.
\newblock Neural discrete representation learning.
\newblock \emph{Advances in neural information processing systems}, 30.

\bibitem[{Vaswani et~al.(2017)Vaswani, Shazeer, Parmar, Uszkoreit, Jones,
  Gomez, Kaiser, and Polosukhin}]{vaswani2017attention}
Ashish Vaswani, Noam Shazeer, Niki Parmar, Jakob Uszkoreit, Llion Jones,
  Aidan~N Gomez, {\L}ukasz Kaiser, and Illia Polosukhin. 2017.
\newblock Attention is all you need.
\newblock \emph{Advances in neural information processing systems}, 30.

\bibitem[{Yu et~al.(2022)Yu, Xu, Koh, Luong, Baid, Wang, Vasudevan, Ku, Yang,
  Ayan et~al.}]{parti}
Jiahui Yu, Yuanzhong Xu, Jing~Yu Koh, Thang Luong, Gunjan Baid, Zirui Wang,
  Vijay Vasudevan, Alexander Ku, Yinfei Yang, Burcu~Karagol Ayan, et~al. 2022.
\newblock Scaling autoregressive models for content-rich text-to-image
  generation.
\newblock \emph{arXiv preprint arXiv:2206.10789}.

\bibitem[{Zeng et~al.(2019)Zeng, Li, and Zhang}]{zeng2019pororogan}
Gangyan Zeng, Zhaohui Li, and Yuan Zhang. 2019.
\newblock Pororogan: an improved story visualization model on pororo-sv
  dataset.
\newblock In \emph{Proceedings of the 2019 3rd International Conference on
  Computer Science and Artificial Intelligence}, pages 155--159.

\bibitem[{Zhang et~al.(2017)Zhang, Xu, Li, Zhang, Wang, Huang, and
  Metaxas}]{zhang2017stackgan}
Han Zhang, Tao Xu, Hongsheng Li, Shaoting Zhang, Xiaogang Wang, Xiaolei Huang,
  and Dimitris~N Metaxas. 2017.
\newblock Stackgan: Text to photo-realistic image synthesis with stacked
  generative adversarial networks.
\newblock In \emph{Proceedings of the IEEE international conference on computer
  vision}, pages 5907--5915.

\end{thebibliography}
\bibliographystyle{acl_natbib}

\appendix

\section{PororoSV Cases}
\label{sec:appendix}
\textit{\footnotesize{
\textbf{Case 1:}\\
1. Tongtong opens the door. Crong is now on the Pororo's car. They are entering the Tongtong's house. Tongtong tries to find where the magic wand is.\\
2. The pink magic wand is located behind the chair. Then the magic wand becomes to come out.\\
3. Tongtong finally finds out the magic wands.\\
4. As Tongtong finds out the magic wand Tongtong is confident to make Pororo normal. Tongtong says that Pororo will turn back to normal.\\
5. Pororo jumps high after Tongtong's promise. Tongtong asks Pororo not to mess up.\\
\textbf{Case 2:}\\
1. Pororo and Crong is in Pororo's house. They are standing next to the bed. Pororo is pointing Crong. Crong looks sad.\\
2. Pororo and Crong is in Pororo's house standing next to the bed. Pororo is pointing drawer and Crong looks at it with a sad face.\\
3. Poby is in Pororo's house. Poby is approaching a drawer. Above the drawer there is a book which is slightly open.\\
4. Poby is in Pororo's house searching for something. Poby leans his head down to take a look.\\
5. Poby is in Pororo's house. Poby is thinking something standing still putting his right hand on his jaw.\\
\textbf{Case 3:}\\
1. Pororo dust the snow off from Poby.\\
2. Petty is holding the green block.\\
3. Harry explains situation. Pororo walks toward chair.\\
4. Pororo sits and joins play.\\
5. Eddy and Crong are yelling.\\
\textbf{Case 4:}\\
1. Poby looks at Harry and Harry is talking to Poby while sitting on Poby's shoulder.\\
2. Petty talks and smiles and mops the floor.\\
3. Petty is smiling and mobbing the floor.\\
4. Petty smiles and puts the stuff on the table.\\
5. Poby talks and opens Poby mouth.\\
\textbf{Case 5:}\\
1. Pororo says it was so stinky.\\
2. feeling embarrassed Poby waved Poby hands.\\
3. Pororo thinks who farted just before.\\
4. everyone saw Crong pinching everyone noses.\\
5. Crong is sitting on the toilet.\\
\textbf{Case 6:}\\
1. Poby is leaving Loopy house.\\
2. Poby says bye to Loopy.\\
3. Poby thinks Pororo might have fixed broken chair.\\
4. Poby smiles. Loopy walks toward the chair.\\
5. Loopy is satisfied with fixed chair.\\
\textbf{Case 7:}\\
1. Poby is tired so Poby says to Harry that Poby wants to go to bed with sleepy eyes.\\
2. Harry is surprised. Harry looks at the window\\
3. there are two cactus on the shelf. outside the window is dark already.\\
4. light is turned off and Poby and Harry finish ready to sleep. Harry say good night to Poby.\\
5. light is turned off and Poby and Harry finish ready to sleep. Poby lays down on the bed.\\
\textbf{Case 8:}\\
1. Poby notices that someone is skiing down.\\
2. Pororo is skiing away. Loopy is chasing.\\
3. Pororo notices that Poby is waiting for Pororo.\\
4. Loopy and Poby are lying down. Eddy approaches.\\
5. Poby and Loopy stands up.\\
\textbf{Case 9:}\\
1. Poby comes out of the house and find out his friends.\\
2. Poby explains to the friends that Poby must have fallen asleep inside.\\
3. Eddy is happy to see that Poby is also already at Eddy's house along with other friends.\\
4. Eddy is inviting his friends to come into his house. everyone follows him.\\
5. Eddy is leading his friends into his house. everyone is getting inside.\\
\textbf{Case 10:}\\
1. Pororo and friends came running toward Poby. Poby is watching them.\\
2. Pororo and friends are talking to Poby. Poby has no idea why his friends are acting like this.\\
3. Harry sat on Poby's head. Harry is saying sorry to Poby. Poby looks surprised. Harry looks sad.\\
4. Harry is feeling guilty. meanwhile Poby has no idea what Harry is talking about. Harry is sitting on Poby's head.\\
5. Harry is feeling very sorry to Poby. Harry is talking to Poby on Poby's head.\\
\textbf{Case 11:}\\
1. Poby is brushing pole to Poby's noses for making Poby itchy. Pororo tries sneezing Poby to take out of the air.\\
2. Pororo shows that small helicopter is still working properly.\\
3. Pororo continually suggests Poby trying to sneeze again.\\
4. Poby tries to sneeze to get out of air from his body. However sneezing with Poby's free will is really difficult.\\
5. Poby tries to sneeze to get out of air from his body. However sneezing with Poby's free will is really difficult. Poby gives up sneezing and says Poby can't sneeze anymore.\\
\textbf{Case 12:}\\
1. Poby was keep singing. Suddenly Poby falls down.\\
2. Poby feels ashamed and wants that nobody saw him falling down.\\
3. Seeing Poby through the telescope Eddy secretly smiles and talks to himself that Eddy saw Poby falling down.\\
4. Eddy is interested in seeing things and friends through telescope. Eddy brings telescope and goes to the mountain to observe his friends more.\\
5. Up on the mountain Eddy chooses a target. It is Pororo. Eddy looks through the telescope.\\
}}

\begin{figure*}[h]
 \centering
  \includegraphics[width=\linewidth]{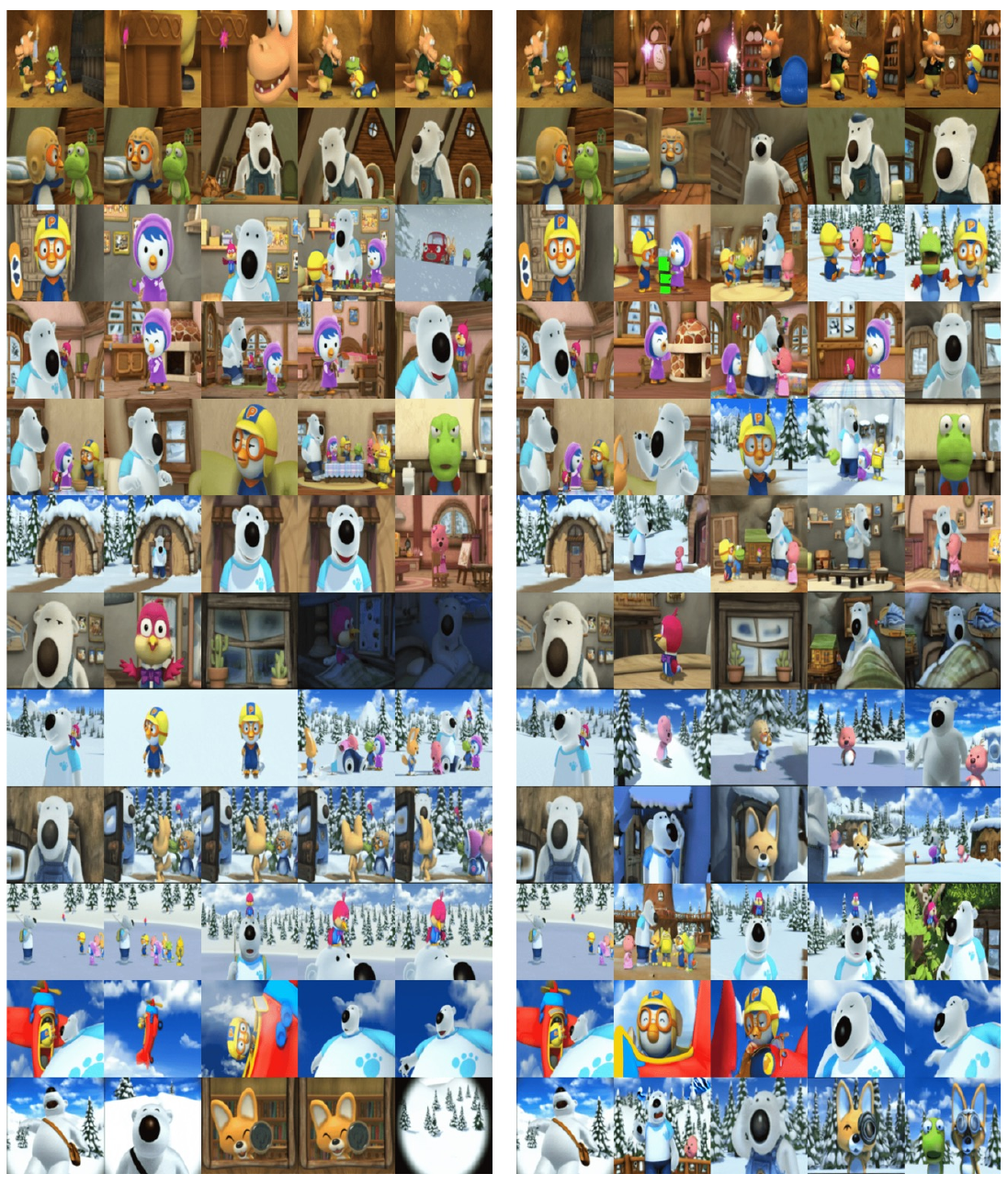}
  \caption{Example of ground truths (left 5 frames) and corresponding generated visual stories  (right 5 frames) on PororoSV. These cases are under story continuation setting.}
  \label{fig:pororo_cases}
\end{figure*}

\section{FlintstonesSV Cases}
\textit{\footnotesize{
\textbf{Case 1:}\\
1.fred and barney stand outside holding blue lunch boxes.fred talks to barney\\
2.Fred stands in the kitchen having a friendly conversation with someone.\\
3.Fred is standing in a room. He is speaking while looking over his shoulder and smiling.\\
4.Fred is trying to kiss Wilma in a room. Wilma is holding a type of plant in her hand.\\
5.Wilma is in a living room adjusting the leafs on a house plant that is sitting on a table while talking then she stands up straight and turns her head.\\
\textbf{Case 2:}\\
1.A Lounging creature is lounging around the room and talking.\\
2.Wilma is ironing a shirt in the laundry room.\\
3.Wilma is in the living room. Wilma is ironing. Wilma is bobbing her head. \\
4.Wilma is in a room. She talks to someone.\\
5.Wilma is in a room. She is talking.\\
\textbf{Case 3:}\\
1.Fred sits in the living room and speaks to Barney, who waits to respond.\\
2.Barney stands and talks with someone in the living room.\\
3.Fred and Barney are in the quarry. Fred is speaking to Barney.\\
4.Fred and Barney look worried. Fred and Barney are behind the wall in the yard. Barney and Fred are talking to each other. \\
5.Barney is talking to Fred outside behind a stone fence. Fred begins to slump down and look sad.\\
\textbf{Case 4:}\\
1.Fred is riding in the car thinking and talking to himself.\\
2.Fred touches his chin then crosses his arm while outside.\\
3.Fred is walking outside while speaking out loud.\\
4.Fred and Barney are riding in the car with golf clubs strapped to the bumper.\\
5.Fred is standing in the room, talking to someone off screen left.\\
\textbf{Case 5:}\\
1.Fred and Barney are standing outside next to the stone wall. Barney is wearing an outfit that makes him look like a boy scout. Fred says something to Barney and then points at him.\\
2.Fred and Barney are standing on a sidewalk. Barney is speaking to Fred, while Fred listens silently with his hands on his hips.\\
3.Barney is outside talking.\\
4.Fred and Barney stand in the yard. They speak to each other. \\
5.Fred and barney are standing outside talking.  They are in front of the wall and barney has a hate on.\\
\textbf{Case 6:}\\
1.Fred sits in the living room and speaks to Barney, who waits to respond.\\
2.Barney stands and talks with someone in the living room.\\
3.Fred and Barney are in the quarry. Fred is speaking to Barney.\\
4.Fred and Barney look worried. Fred and Barney are behind the wall in the yard. Barney and Fred are talking to each other. \\
5.Barney is talking to Fred outside behind a stone fence. Fred begins to slump down and look sad.\\
\textbf{Case 7:}\\
1.Wilma is in the room, she is talking.\\
2.Wilma is in the dining room talking to someone then she starts to laugh.\\
3.Barney slides towards doorway, and opens door.\\
4.Wilma is sitting in the dining room at the table while talking on the phone.\\
5.Wilma is in the dining room. She sits at the table on the phone. Fred is wheeled into the room laying in a bed. As Fred enters, Wilma lowers the phone and looks at him with concern and surprise.\\
\textbf{Case 8:}\\
1.Fred is in a living room kneeling next to a blue chair.\\
2.Pebbles and Fred are standing in a room talking.  Pebbles turns her head and Fred shrugs his shoulders. \\
3.Fred and wilma talk in the bedroom, fred laughs in response to what Wilma says.\\
4.Fred makes an angry comment while sitting in the room.\\
5.Fred stands in the kitchen with an ice block on his head. Then someone reaches up and pats the ice. Fred makes a face and the ice starts melting.\\
\textbf{Case 9:}\\
1.Barney is talking in the living room.\\
2.Betty is standing in a room, hanging up balloons. She is talking to someone, as she hangs up a green balloon. \\
3.Betty is in a room. She stands on a stool and holds a balloon in one hand while talking to someone on the ground. The room is decorated for a party.\\
4.Wilma is in the living room.  Wilma is talking.  \\
5.There is a bird that has its head lowered in the room .\\
\textbf{Case 10:}\\
1.Wilma and Betty are in the room. They are talking to one another while standing.\\
2.Betty and Wilma are standing in a room. Wilma has bones in her hair. Betty is talking to Wilma.\\
3.Wilma is wearing a bone curler in her hair while Betty talks to her in a room.\\
4.Betty and wilma are talking in a room.\\
5.Wilma and Betty are standing in a room by the window. They keep looking out the window while Wilma holds the curtain.\\
\textbf{Case 11:}\\
1.Mr slate is driving his car and laughing.\\
2.A police officer in a police station sits at a desk and talks into a speaker while looking at a stack of papers. He turns the speaker away from his mouth.\\
3.A Small Policeman behind wheel and a Policeman with Brown Mustache sit in their police car blinking.\\
4.The officer that is driving the car is speaking to the officer with mustache.\\
5.Fred and Barney talk as they sit in the car.\\
\textbf{Case 12:}\\
1.Barney is sitting outside in a chair reading out loud.\\
2.Barney is reading the news papers in the backyard.\\
3.The scene begins with no one in the picture. Barney emerges from hiding behind a stone wall that is in front of him. He is standing outside in the yard. The house is to his right. He says something and then points to himself with his thumb.\\
4.Barney is outside.  He is sitting on a stone wall and is talking.\\
5.Fred and Barney are in the yard. Fred is yelling at Barney. Fred is holding Barney with a fist raised.\\
}}

\begin{figure*}[h]
\centering
  \includegraphics[width=\linewidth]{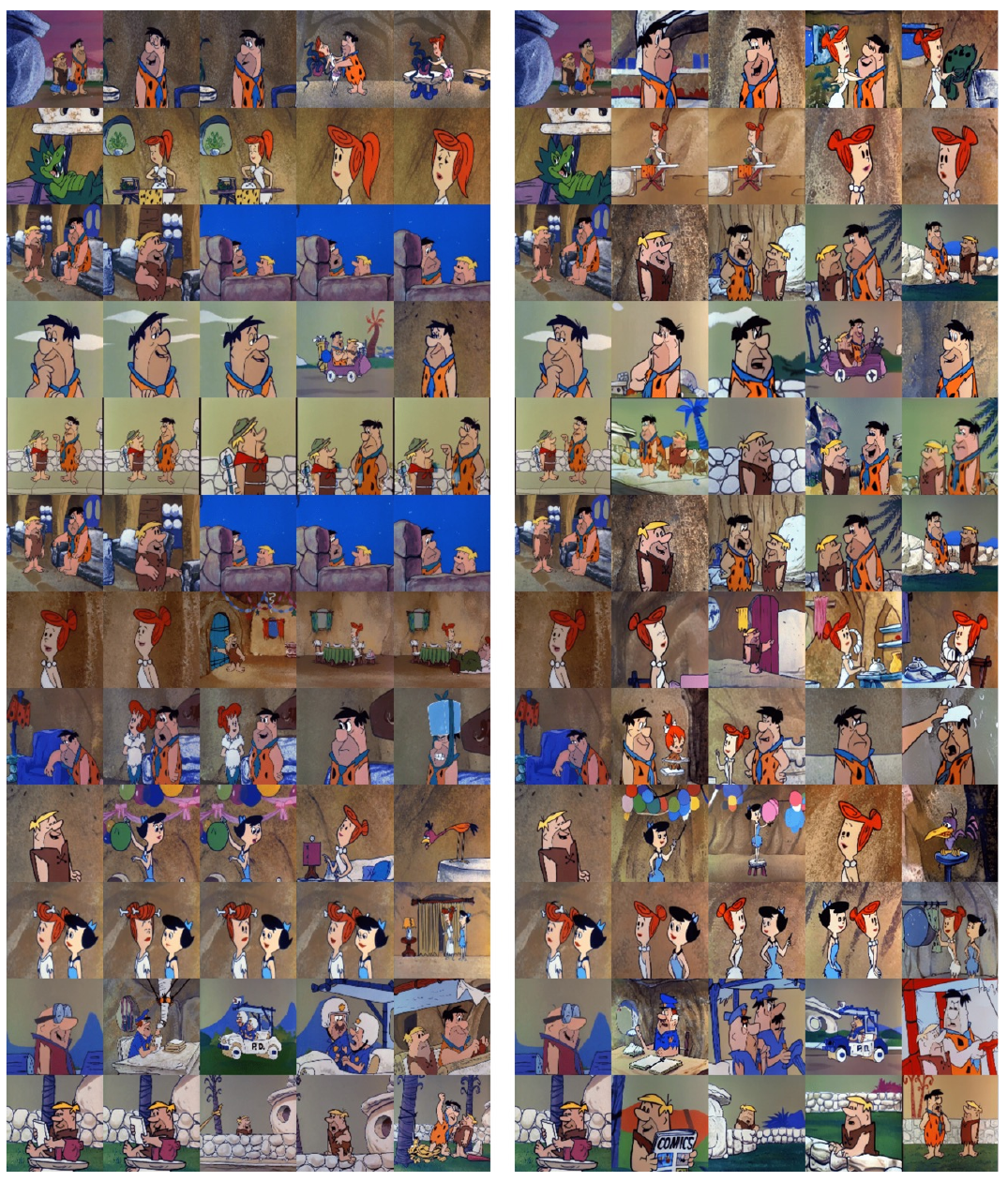}
  \caption{Example of ground truths (left 5 frames) and corresponding generated visual stories  (right 5 frames) on FlintstonesSV. These cases are under story continuation setting.}
  \label{fig:flint_cases}
\end{figure*}

\end{document}